# A Reconfigurable Low Power High Throughput Architecture for Deep Network Training

Raqibul Hasan, and Tarek M. Taha
Department of Electrical and Computer Engineering
University of Dayton, Ohio
Email: {hasanm1, tarek.taha}@udayton.edu

*Abstract*—General purpose computing systems are used for a large variety of applications. Extensive supports for flexibility in these systems limit their energy efficiencies. Neural networks, including deep networks, are widely used for signal processing and pattern recognition applications. In this paper we propose a multicore architecture for deep neural network based processing. Memristor crossbars are utilized to provide low power high throughput execution of neural networks. The system has both training and recognition (evaluation of new input) capabilities. The proposed system could be used for classification, dimensionality reduction, feature extraction, and anomaly detection applications. The system level area and power benefits of the specialized architecture is compared with the NVIDIA Telsa K20 GPGPU. Our experimental evaluations show that the proposed architecture can provide up to five orders of magnitude more energy efficiency over GPGPUs for deep neural network processing.

## I. INTRODUCTION

Reliability and power consumption are among the main obstacles to continued performance improvement of future multicore computing systems [1]. As a result, several research groups are investigating the design of energy efficient processors from different aspects. These include architectures for approximate computation utilizing dynamic voltage scaling technique, dynamic precision control, and inexact hardware [2,3]. Emerging non-volatile memory technologies are being investigated as low power on-chip caches [4]. Application specific architectures are also proposed for several application domains such as signal processing and video processing.

Interest in specialized architectures for accelerating neural networks has increased significantly because of their ability to reduce power, increase performance, and allow fault tolerant computing. Recently IBM has developed the TrueNorth chip [5] consisting of 4,096 neurosynaptic cores interconnected via an intra-chip network. Their synapse element is SRAM based and off-chip training is utilized. DaDianNao [6] is an accelerator for deep neural network (DNN) and convolutional neural network (CNN). In this system, neuron synaptic weights are stored in eDRAM and later brought into Neural Functional Unit for execution.

Recently deep neural networks (or deep networks) have gained significant attention because of their superior performance for classification and recognition applications. Training and evaluation of a deep network are both computationally and data intensive tasks. This paper presents a generic multicore architecture for training and recognition of deep network applications. The system has both unsupervised and supervised learning capabilities. The proposed system could be used for classification, unsupervised clustering, dimensionality reduction, feature extraction and anomaly detection applications.

Memristor [7] is a novel non-volatile device having a large varying resistance range. Physical memristors can be laid out in a high density grid known as a crossbar [8]. A memristor crossbar can evaluate many multiply-add operations in parallel in analog domain which are the dominant operations in neural networks. We are using memristor crossbars in the proposed system which provide high synaptic weight density and parallel analog processing consuming very low energy. In this system processing happens at physical location of the data. Thus data transfer energy and functional unit energy consumptions are saved significantly.

Both the training and the recognition phases of the neural networks were examined. As deep networks deal with large networks, efficient approaches to simulate and implement large memristor crossbars for these networks are important. We have presented a novel method to accurately simulate large crossbars at high speed. Detailed circuit level simulations of memristor crossbars were carried out to verify the neural operations. We have evaluated the power, area, and performance of the proposed multicore system and compared them with a GPU based system. Our results indicate that the memristor based architecture can provide up to five orders of magnitude more energy efficiency over GPU for the selected benchmarks.

The related memristor core design works in this area are [9,10] where the impact on area, power, and throughput are examined for systems that carry out recognition tasks only. Unsupervised training or deep network training is not examined in these studies. These systems are based on ex-situ training and do not examine on-chip training and corresponding energy consumptions.

The rest of the paper is organized as follows: section II demonstrates the overall multicore architecture. Section III gives background on deep networks and their training methods. Section IV describes neuron circuit design, memristor based neural network and training circuit design. Section V describes large memristor crossbar simulation approach and memristor neural core design. Sections VI and VII describe experimental setup and results respectively. Section VIII describes related works in the area and finally in section IX we summarize our work.

## II. SYSTEM OVERVIEW

Multicore architectures are widely used to exploit task level parallelism. We assumed a multicore neural architecture, for neural network applications as shown in Fig. 1. Memristor crossbar neural cores are utilized to provide low power, high throughput execution of neural networks. The cores in the system are connected through an on-chip routing network (R). The neural cores receive their inputs from main memory holding the training data through the on-chip routers and a buffer between the routing network and the main memory as shown in Fig. 1.

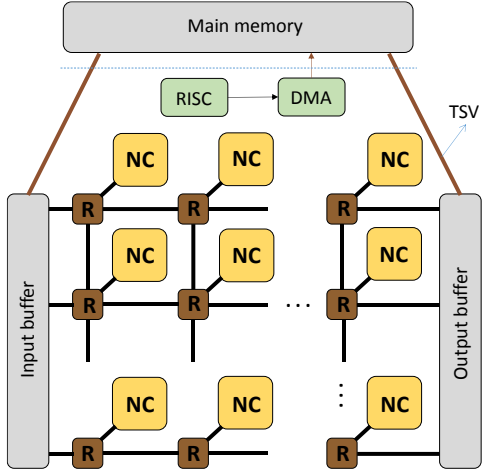

Fig. 1. Proposed multicore system with several neural cores (NC) connected through a 2-D mesh routing network. (R: router).

Since the proposed architecture is geared towards machine learning applications, the training data will be used multiple times during training. As a result, access to a high bandwidth memory system is needed, and thus we propose the use of a 3D stacked DRAM. Using through Silicon vias reduces memory access time and energy, thereby reduces overall system power. To allow efficient access to the main memory a DMA controller is utilized that is initialized by a RISC processing core. A single issue pipelined RISC core is used to reduce power consumption. Another possible interface to the proposed architecture could be with a 3D stacked sensor chip. Input data coming from the sensor chip will be processed in the neural chip in real time.

An on-chip routing network is needed to transfer neuron outputs among cores in a multicore system. In feed-forward neural networks, the outputs of a neuron layer are sent to the following layer after every iteration (as opposed to a spiking network, where outputs are sent only if a neuron fires). This means that the communication between neurons is deterministic and hence a static routing network can be used for the core-to-core communications. In this study, we assumed a static routing network as this would be lower power consuming than a dynamic routing network.

SRAM based static routing is utilized to facilitate re-programmability in the switches. Fig. 2 shows the routing switch design. Note carefully that the switch allows outputs from a core to be routed back into the core to implement recurrent networks or multi-layer neural networks where all the neuron layers are within the same core.

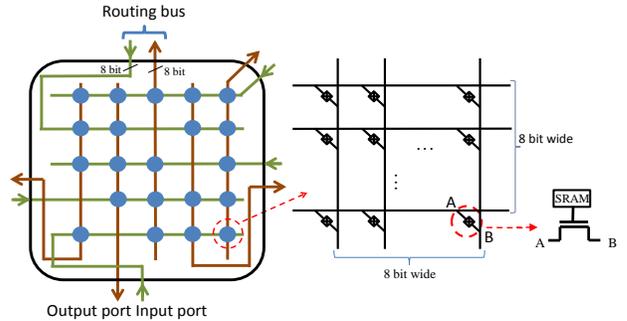

Fig. 2. SRAM based static routing switch. Each blue circle in the left part of the figure represents the 8x8 SRAM based switch shown in the middle (assuming a 8-bit network bus).

## III. DEEP NEURAL NETWORKS

### A. Deep Networks

Deep neural networks have become highly popular recently for classification and recognition applications. Fig. 3 shows a block diagram of a deep network. There is strong similarity between the architectures of multi-layer neural networks and deep networks. A deep neural network has multiple hidden layers and large number of neurons in a layer which can learn significantly more complex functions.

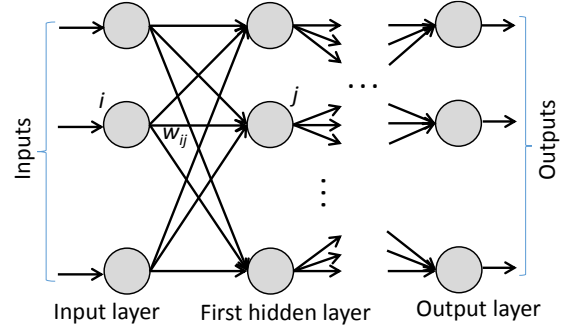

Fig. 3. Block diagram of a deep network.

Neurons are the building blocks of a deep network. A neuron in a deep network performs two types of operations: (i) a dot product of the inputs $x_1,\ldots,x_n$ and the weights $w_1,\ldots,w_n$, and (ii) the evaluation of an activation function. The dot product operation for a neuron $j$ can be seen in Eq. (1). The activation function evaluation is shown in Eq. (2). In a deep network, a nonlinear differentiable activation function is desired (such as $tan^{-1}(x)$).

$$DP_j = \sum_{i=1}^{n} x_i w_{ij} \quad (1)$$
$$y_j = f(DP_j) \quad (2)$$

### B. Training of Deep Networks

The training of deep networks is different from the training of multi-layer neural networks. Unlike multi-layer neural networks, a deep network does not perform well if it is trained using only a supervised learning algorithm on the entire network. Since deep networks have many layers of neurons, these networks are typically trained in a two step process: an unsupervised layer-wise pre-training step followed by a supervised training step of the entire network [11]. In this paper we utilized autoencoders for layer-wise unsupervised pre-training for memristor based deep networks. We utilized back-propagation based training circuit for both the layer-wise pre-training and the supervised fine tuning of the whole network.

The architecture of an autoencoder is similar to a multi-layer neural network, as shown in Fig. 4. An autoencoder tries to learn a function $h_{W,b}(x) \approx x$. That is, it tries to learn an approximation to the identity function, such that the network's output $x'$ is similar to the input $x$. By placing constraints on the network, such as by limiting the number of hidden units, we can discover useful patterns within the data. Gradient descent training is generally utilized for training an autoencoder.

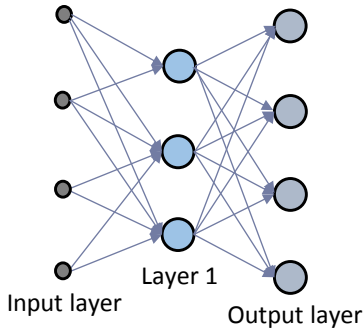

Fig. 4. Two layer network having four inputs, three hidden neurons and four output neurons.

## IV. MEMRISTOR NEURON CIRCUIT AND NEURAL NETWORK TRAINING

### A. Neuron Circuit

In this paper we have utilized memristors as neuron synaptic weights. The circuit in Fig. 5(a) shows the memristor based neuron circuit design. In this circuit, each data input is connected to two virtually grounded op-amps (operational amplifiers) through a pair of memristors. For a given row, if the conductance of the memristor connected to the first column ($\sigma_A^+$) is higher than the conductance of the memristor connected to the second column ($\sigma_A^-$), then the pair of memristors represents a positive synaptic weight. In the inverse situation, when $\sigma_A^+ < \sigma_A^-$, the memristor pair represents a negative synaptic weight.

In Fig. 5(a) currents through the first and second columns are $A\sigma_{A+} + \cdots + \beta\sigma_{\beta+}$ and $A\sigma_{A-} + \cdots + \beta\sigma_{\beta-}$ respectively. The output of the op-amp, connected directly with the second column, represents the neuron output. In the non-saturating region of the second op-amp, the output $y_j$ of the neuron circuit is given by

$$y_j = R_f[\{A\sigma_{A+} + \cdots + \beta\sigma_{\beta+}\} - \{A\sigma_{A-} + \cdots + \beta\sigma_{\beta-}\}]$$
$$= R_f[A(\sigma_{A+} - \sigma_{A-}) + \cdots + \beta(\sigma_{\beta+} - \sigma_{\beta-})]$$

Assume that
$$DP_j = 4R_f[A(\sigma_{A+} - \sigma_{A-}) + \cdots + \beta(\sigma_{\beta+} - \sigma_{\beta-})]$$
(here $4R_f$ is a constant)

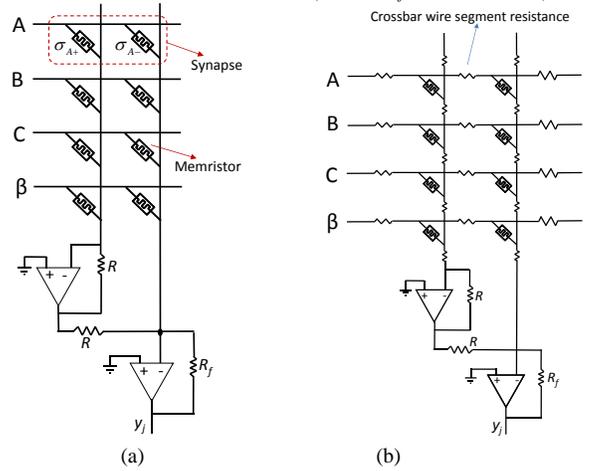

Fig. 5. Memristor based neuron circuit. *A, B, C* are the inputs and $y_j$ is the output.

When the power rails of the op-amps, $V_{DD}$ and $V_{SS}$ are set to *0.5V* and *-0.5V* respectively, the neuron circuit implements the activation function $h(x)$ as in Eq. (3) where $x = 4R_f[A(\sigma_{A+} - \sigma_{A-}) + \cdots + \beta(\sigma_{\beta+} - \sigma_{\beta-})]$. This implies, the neuron output can be expressed as $h(DP_j)$.

$$h(x) = \begin{cases} 0.5 & if\ x > 2 \\ \dfrac{x}{4} & if\ |x| < 2 \\ -0.5 & if\ x < -2 \end{cases} \quad (3)$$

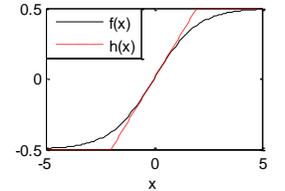

Fig. 6. Plot of functions *f(x)* and *h(x)* which show *h(x)* is approximating *f(x)* closely.

Fig. 6 shows that *h(x)* closely approximates the sigmoid activation function, $f(x) = \dfrac{1}{1+e^{-x}} - 0.5$. The values of $V_{DD}$ and $V_{SS}$ are chosen such that no memristor gets a voltage greater than $V_{th}$ across it during evaluation. Our experimental evaluations consider memristor crossbar wire resistance. The schematic of a memristor based neuron circuit considering wire resistance is shown in Fig. 5(b).

### B. Memristor Based Neural Network Implementation

Recall that the structures of a multi-layer neural network and an autoencoder are similar. Both can be viewed as a feed forward neural network. Fig. 4 shows a simple two layer feed forward neural network with four inputs, four outputs, and three hidden layer neurons. Fig. 7 shows a memristor crossbar based circuit that can be used to evaluate the neural network in Fig. 4. There are two memristor crossbars in this circuit, each representing a layer of neurons. Each crossbar utilizes the neuron circuit shown in Fig. 5(a).

In Fig. 7, the first layer of the neurons is implemented using an 5×6 crossbar. The second layer of two neurons is implemented using a 4×8 memristor crossbar, where 3 of the inputs are coming from the 3 neuron outputs of the first crossbar. One additional input is used for bias. Applying the inputs to a crossbar, an entire layer of neurons is processed in parallel using Ohm's law in the analog domain (in one step). No multiplier or adder is needed in this processes.

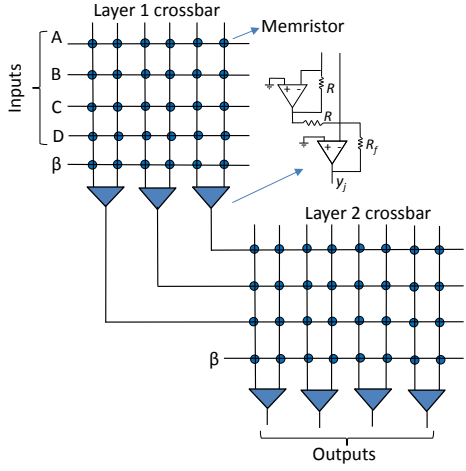

Fig. 7. Schematic of the neural network shown in Fig. 4 for forward pass. Each crossbar implements a layer of neurons.

*C. The Training Algorithm*

Both the layer-wise unsupervised pre-training (using autoencoder) and the supervised training, utilized in a deep network, are built on the back-propagation (BP) algorithm [12]. In this paper we utilized the stochastic BP algorithm, where the network weights are updated after each input is applied. For autoencoders, the inputs to the autoencoder are used as the targets for the final layer of the autoencoder. The training algorithm utilized in this paper is described below:

1) Initialize the memristors with high random resistances.
2) For each input pattern *x*:
   i) Apply the input pattern *x* to the crossbar circuit and evaluate the $DP_j$ values and outputs ($y_j$) of all neurons (hidden neurons and output neurons).
   ii) For each output layer neuron *j*, calculate the error, $\delta_j$, based on the neuron output ($y_j$) and the target output ($t_j$). Here *f* is the neuron activation function.
   $$\delta_j = (t_j - y_j)f'(DP_j) \quad (4)$$
   iii) Back propagate the errors for each hidden layer neuron *j*.
   $$\delta_j = (\sum_k \delta_k w_{k,j}) \times f'(DP_j) \quad (5)$$
   where neuron *k* is connected to the previous layer neuron *j*.
   iv) Determine the amount, $\Delta w$, that each neuron's synapses should be changed (*2η* is the learning rate):
   $$\Delta w_j = 2\eta \times \delta_j \times x \quad (6)$$
3) If the error in the output layer has not converged to a sufficiently small value, goto step 2.

*D. Circuit Implementation of the Back-propagation Training Algorithm*

The implementation of the training circuit can be broken down into the following major steps:
1. Apply inputs to layer 1 and record the layer 2 neuron outputs and errors.
2. Back-propagate layer 2 errors through the second layer weights and record the layer 1 errors.
3. Update the synaptic weights.

The circuit implementations of these steps are detailed below:

***Step 1:*** A set of inputs is applied to the layer 1 neurons, and both layer 1, and layer 2 neurons are processed. In Eq. (4 and 5) we need to evaluate the derivative of the activation function for the dot product of the neuron inputs and weights ($DP_j$). The $DP_j$ value of neuron *j* is essentially the difference of the currents through the two columns implementing the neuron (this can be approximated based on $y_j$ in Fig. 5(a)). The $DP_j$ value of each neuron *j* is discretized and $f'(DP_j)$ is evaluated using a lookup table. The $f'(DP_j)$ value of each neuron is stored in a buffer. The layer 2 neuron errors are evaluated based on the neuron outputs ($y_j$), the corresponding targets ($t_j$) and $f'(DP_j)$ as shown in Fig. 8. First ($t_j$-$y_j$) is evaluated and discretized using an ADC (analog to digital converter). Then ($t_j$-$y_j$), and $f'(DP_j)$ are multiplied using a digital multiplier and the evaluated $\delta_j$ value is stored in a register.

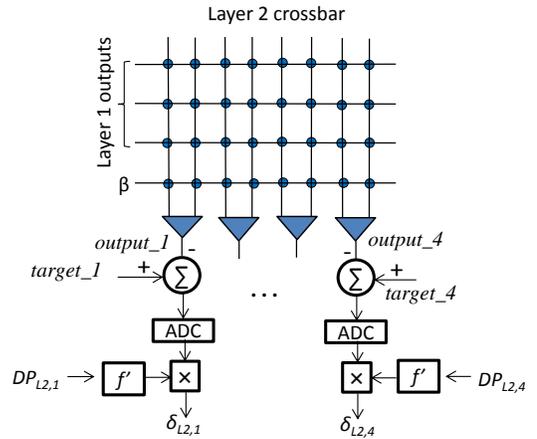

Fig. 8. Output layer error generation circuits which take neuron outputs, corresponding targets, and $DP_j$ values as input.

***Step 2:*** The layer 2 errors ($\delta_{L2,1}$,.., $\delta_{L2,4}$) are applied to the layer 2 weights after conversion from digital to analog form as shown in Fig. 9 to generate the layer 1 errors ($\delta_{L1,1}$ to $\delta_{L1,3}$). The memristor crossbar in Fig. 9 is the same as the layer 2 crossbar in Fig. 7. Assume that the synaptic weight associated with input *i*, neuron *j* (second layer neuron) is $w_{ij}=\sigma_{ij}^+ - \sigma_{ij}^-$ for *i=1,2,3* and *j=1,2,..,4*. In the backward phase, we want to evaluate the layer 1 error

$$\delta_{L1,i} = (\Sigma_j w_{ij} \delta_{L2,j}) f'(DP_{L1,i}) \quad \text{for } i=1,2,3 \text{ and } j=1,2,..,4.$$
$$= (\Sigma_j (\sigma_{ij}^+ - \sigma_{ij}^-) \delta_{L2,j}) f'(DP_{L1,i})$$
$$= (\Sigma_j \sigma_{ij}^+ \delta_{L2,j} - \Sigma_j \sigma_{ij}^- \delta_{L2,j}) f'(DP_{L1,i}) \quad (7)$$

The circuit in Fig. 9 is essentially evaluating the same operations as Eq. (7), applying both $\delta_{L2,j}$ and $-\delta_{L2,j}$ to the crossbar columns for $j=1,2,..,4$. The back propagated (layer 1) errors are stored in buffers for updating crossbar weights in step 3. To reduce the training circuit overhead, we can multiplex the back propagated error generation circuit as shown in Fig. 10. In this circuit, by enabling the appropriate pass transistor, back propagated errors are sequentially generated and stored in buffers. Access to the pass transistors is controlled by a shift register. Same multiplexing approach could also be used for the layer 2 error generation in step 1. In this approach the time complexity of the back propagation step will be $O(m)$ where $m$ is the number of inputs in a layer of neurons.

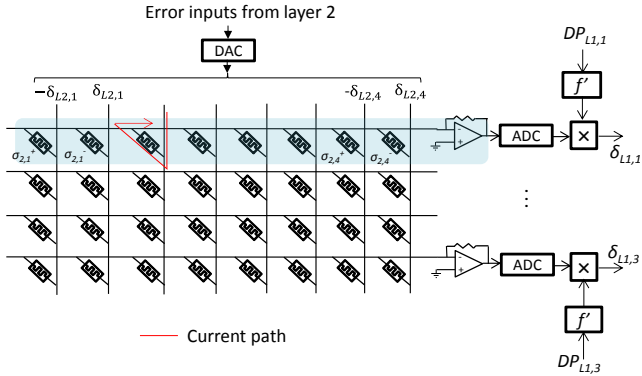

Fig. 9. Schematic of the neural network shown in Fig. 4 for back propagating errors to layer 1.

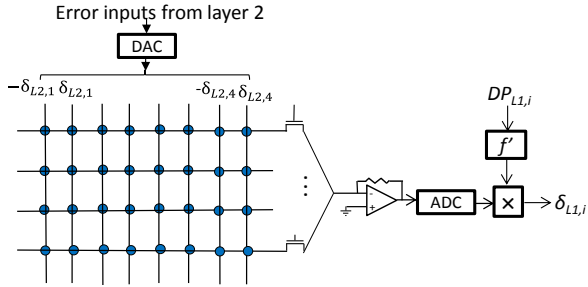

Fig. 10. Implementing back propagation phase multiplexing error generation circuit.

***Step 3:*** Weight update procedure in the memristor crossbar is similar to [13]. Main difference from the design in [13] is that we are utilizing two memristors per synapse while [13] used single memristor per synapse. Two memristors per synapse design has two times more synaptic weight precision than a single memristor per synapse design. For training, pulses of variable amplitude and variable duration are produced. The amplitude of the training signal is modulated by input $x_i$ and the duration of the training pulse is modulated by $\eta \times \delta_j$. The combined effect of the two voltages applied across the memristor will update the conductance by an amount proportional to $\eta \times \delta_j \times x_i$.

## V. NEURAL CORE DESIGN AND LARGE MEMRISTOR CROSSBAR SIMULATION

### A. Memristor Neural Core

We considered memristor crossbar wire segment resistance 1.5 ohms and resistance of the trained memristors were, on average, between 1M ohms to 10M ohms. We examined training of neural networks using crossbars of large sizes. Large crossbars (size over 400×200), make training challenging due to sneak current paths. The neural networks implemented using crossbars of sizes 400×200 or smaller were able to learn the desired functionalities smoothly.

Fig. 12 shows the memristor based single neural core architecture. It consists of a memristor crossbar of size 400×200, input and output buffers, a training unit, and a control unit. Our objective was to take as big crossbar as possible, because this enables more computations to be done in parallel. The control unit will manage the input, output buffers and will interact with the specific routing switch associated with the core. The control unit will be implemented as a finite state machine and thus will be of low overhead. Processing in the core is in analog and entire core processes in one cycle for an input.

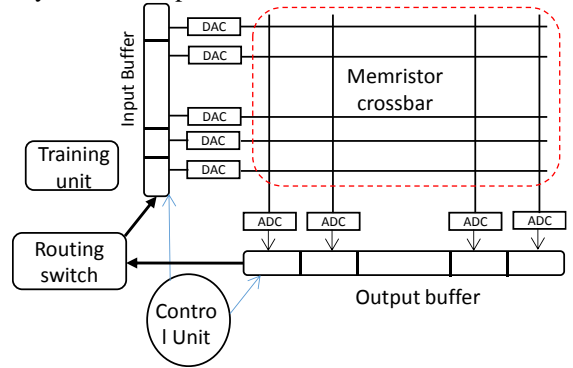

Fig. 12. Single memristor neural core architecture having input buffer, output buffer, training circuits and a control unit.

The neural cores communicate between each other in digital form as it is expensive to exchange and latch analog signals. Neuron outputs are discretized using a three bit ADC converter for each neuron and are stored in the output buffer for routing. Precision of ADC converters is a trade-off between accuracy and circuit area, power overhead. 22% of the core power is consumed by the ADCs (3 bit) in the recognition phase of a neural core. Inputs come to the neural core through the routing network in digital form and are stored in the input buffer. Inputs are applied to the memristor crossbar, converting them into analog form.

### B. Large Crossbar Simulation

In deep networks, the layers can be very wide, having a large number of inputs and outputs. This implies a large memristor crossbar would be needed to implement a layer of neurons. SPICE level accurate simulations of the crossbars are needed to correctly model these sneak-paths. The SPICE simulation of large memristor crossbars are very time consuming (about a day per iteration). We have designed a MATLAB framework for large memristor crossbar simulation which is very fast compared to SPICE simulation and is as accurate as SPICE simulations (less than a minute per iteration).

Consider the *M×N* crossbar in Fig. 11. There are *MN* memristors in this crossbar circuit, *2MN* wire segments, and *M* input drivers. For any given set of crossbar input voltages, we need to determine the *2MN* terminal (node) voltages across the memristors. The Jacobi method of solving systems of linear equations was used to determine these unknown voltages. All the nodes on the crossbar rows were initialized to the applied input voltages, while all the nodes on the crossbar columns were initialized to zero volts. We then repeatedly calculated currents through the crossbar memristors and updated the node voltages until convergence. The node voltages are updated based on the voltage drop across the crossbar wire resistances considering the currents through the memristors. We have compared the crossbar simulation results obtained from SPICE and the MATLAB framework for crossbars of different sizes and observed that the corresponding results match.

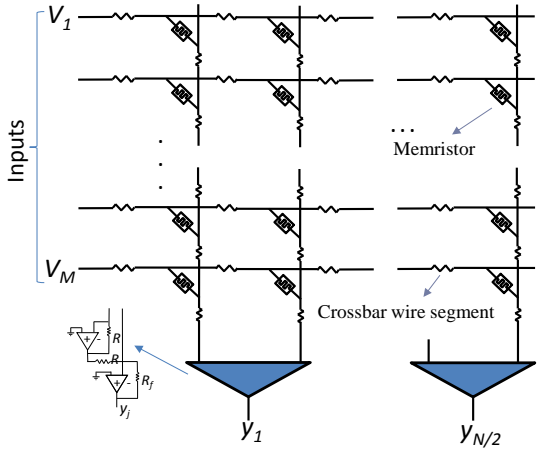

Fig. 11. Schematic of a *M×N* crossbar implementing a layer of *N/2* neurons.

## VI. EXPERIMENTAL SETUP

### A. Applications and Datasets

We have examined the MNIST [14], Caltech-101 [15] and ISOLET [16] datasets for classification tasks using deep networks. Aautoencoder were utilized for unsupervised layer-wise pre-training of the deep networks. A part of the Caltech-101 dataset was utilized for classification which considered only motorcycle and airplane images. The images were resized as 200×300 images. The objective was to demonstrate evaluation for large image data. An autoencoder based anomaly detection was examined on the KDD dataset [17]. Table I shows the neural network configurations for different datasets and applications.

Table I: Neural network configurations.

|   | Dataset | Configuration |
|---|---------|---------------|
| Anomaly detection | KDD | 41→15→41 |
| Classification | MNIST | 784→300→200→100→10 |
|   | ISOLET | 617→2000→1000→500→250→26 |
|   | Caltech | 60,000→800→1 |

### B. Mapping Neural Networks to Cores

The neural hardware are not able to time multiplex neurons as their synaptic weights are stored directly within the neural circuits. Hence a neural network's structure may need to be modified to fit into a neural core. In cases where the networks are significantly smaller than the neural core synaptic array, multiple neural layers are mapped to a core. In this case, the layers execute in a pipelined manner, where the outputs of layer 1 are fed back into layer 2 on the same core through the core's routing switch.

When a software network layer is too large to fit into a core (either because it needed too many inputs or it had too many neurons), the network layer is split amongst multiple cores. Splitting a layer across multiple cores due to a large number of output neurons is trivial. When there are too many inputs per neuron for a core, each neuron is split into multiple smaller neurons as shown in Fig. 13. When splitting a neuron, the network needs to be trained based on the new network topology. As the network topology is determined prior to training (based on the neural hardware architecture), the split neuron weights are trained correctly. This is similar to the mechanism used in convolutional neural networks. This approach is essentially restricting the receptive fields of the neurons [18]. Impact of this restricted neural network mapping approach is shown in Fig. 17.

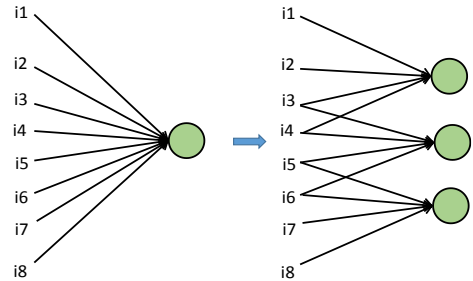

Fig. 13. Splitting a neuron into multiple smaller neurons.

### C. Memristor Device Model

Simulation of the memristor device used an accurate model of the device published in [19]. The memristor device simulated in this paper was published in [20] and the switching characteristics for the model are displayed in Fig. 14. This device was chosen for its high minimum resistance value and large resistance ratio. According to the data presented in [20] this device has a minimum resistance of 10 kΩ, a resistance ratio of $10^3$, and the full resistance range of the device can be switched in 20 μs by applying 2.5 V across the device.

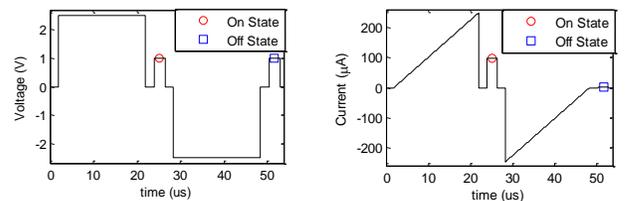

Fig. 14. Simulation results displaying the input voltage and current waveforms for the memristor model [19] that was based on the device in [20]. The following parameter values were used in the model to obtain this result: $V_p$=1.3V, $V_n$=1.3V, $A_p$=5800, $A_n$=5800, $x_p$=0.9995, $x_n$=0.9995, $α_p$=3, $α_n$=3, $a_1$=0.002, $a_2$=0.002, $b$=0.05, $x_0$=0.001.

## D. Area Power Calculations

For the memristor cores, detailed SPICE simulations were used for power and timing calculations of the analog circuits (drivers, crossbar, and activation function circuits). These simulations considered the wire resistance and capacitance within the crossbar as well. The results show that the crossbar required 20 ns to be evaluated. As the memristor crossbars evaluate all neurons in one step, the majority of the time in these systems is spent in transferring neuron outputs between cores through the routing network. We assumed that routing would run at 200 MHz clock resulting in 4 cycles needed for crossbar processing. The routing link power was calculated using Orion [21] (assuming 8 bits per link). Off-chip I/O energy was also considered as described in section II. Data transfer energy via TSV was assumed to be 0.05 pJ/bit [22].

## VII. RESULTS

### A. Unsupervised Training Result

Unsupervised training of a memristor based autoencoder was examined on the MNIST dataset. The network configuration utilized was 784→100→784 (784 inputs, 100 hidden neurons and 784 output neurons). After training (using 1000 digits), the hidden neuron outputs give the feature representation of the corresponding input data in a reduced dimension of 100. Fig. 15 shows the input digits and the corresponding reconstructed digits obtained from the autoencoder (good reconstruction implies good encoding).

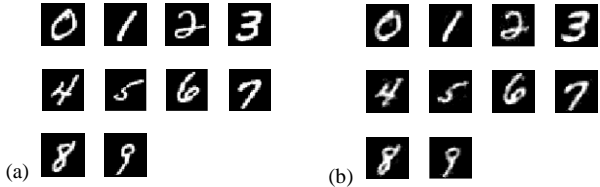

(a)                          (b)

Fig. 15. Test results from the trained autoencoder: (a) input digits (b) corresponding reconstructed digits.

### B. Deep Network Training

Memristor based deep network was examined for the MNIST dataset with configuration 784→200→100→10. To reduce the simulation time, 10000 data were used for training. The network was trained in a two step process: autoencoders were used for layer-wise pre-training, followed by supervised training of the whole network. Fig. 16 shows the supervised training graphs obtained from the MATLAB framework based circuit simulation as well as from the software implementation. The results show that the deep networks were able to learn the desired classification application.

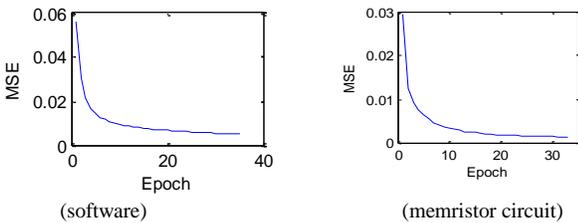

(software)                 (memristor circuit)

Fig. 16. Training results of the memristor based deep neural network. Low mean squared errors (MSE) indicate successful training.

## C. Impact of System Constraints on Application Accuracy

The hardware neural network training circuit differs from the software implementations in the following aspects:
- Limited precision of the discretized neuron errors and $DP_j$ values.
- Limited precision of the discretized neuron output values.
- Each neuron can have a maximum of 400 synapses.

Fig. 17 compares the accuracies obtained from Matlab implementations of the applications in Table I, considering the proposed system constraints and implementations without considering those constraints. It is seen that enforcing the system constraints the applications still give competitive performances.

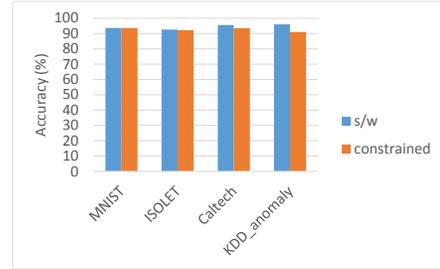

Fig. 17. Impact of memristor system constraints (maximum 400 synapse/neuron, 3 bits neuron output, and 8 bits neuron error precision) on application accuracy.

### D. Single Core Area and Power

The memristor neural core configuration is 400×100, i.e. it can take a maximum of 400 inputs and can process a maximum of 100 neurons. The area of a single memristor neural core is 0.0163 mm$^2$. ). A memristor based synapse is about 146 times denser than a traditional digital SRAM synapse (assuming 8 bits precision in both cases). A 100 mm$^2$ chip, designed based on the proposed approach, would be able to accommodate about 200 million synapses. Table II shows a single memristor core power and timing at different steps of execution. The RISC core is considered to be used only for configuring the cores, routers, and DMA engine. As a result, we assume that the RISC core is turned off afterwards during the actual training or evaluation phases. In the recognition phase, the memristor crossbar consumes 41% and the ADCs (each of 3 bits) consume 22% of the core power while peripheral circuitry consume rest of the power.

Table II: Memristor core timing and power for different execution steps.

|  | Time (us) | Power (mW) |
| --- | --- | --- |
| Forward pass (recognition) | 0.27 | 0.794 |
| Backward pass | 0.80 | 0.706 |
| Weight update | 1.00 | 6.513 |
| Control unit |  | 0.0004 |

### E. System Level Evaluations

**Total system area:** The whole multicore system includes 576 memristor neural cores, one RISC core, one DMA controller, 4 kB of input buffer and 1 kB of output buffer. The RISC core area was evaluated using McPat [23] and came to 0.52 mm². The total system area was 9.94 mm².

**Energy efficiency:** We have compared the throughput and energy benefits of the proposed architecture over a Nvidia Tesla K20 GPU. The system consumes 225 W power. The area of the GPU is 561 mm² using a 28 nm process. We made sure that the GPU implementations are as efficient as possible. The number of threads are enough to keep all the cores busy. The maximum number of threads launched for a CUDA kernel was 2000*1000. Table III shows the training time for single iteration. We ran the applications on GPU for more than 1000 iterations and determined single iteration time from the total training time. The training times do not include the time for copying data from host the memory to the device memory and vice versa.

Figures 18 and 19 show the throughput and energy efficiencies respectively of the proposed system over GPU for different applications during training. For training the proposed architecture provides up to 9.9× speedup and up to five orders of magnitude more energy efficiency over GPU.

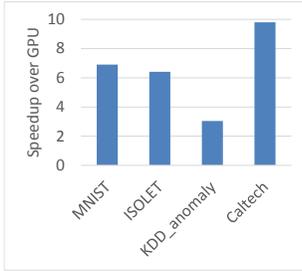
Fig. 18. Application speedup over GPU for training.

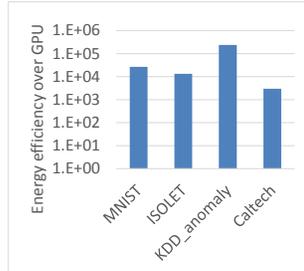
Fig. 19. Energy efficiency of the proposed system over GPU for training.

Figures 20 and 21 show the throughput and energy efficiencies of the proposed system over GPU for different applications respectively during evaluation of new inputs. For recognition, the proposed architecture provides up to 50× speedup and five to five orders of magnitude more energy efficiency over the GPU.

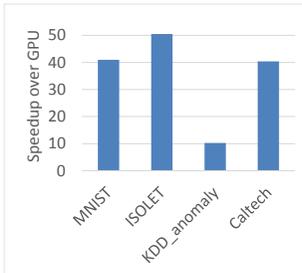
Fig. 20. Application speedup over GPU for recognition.

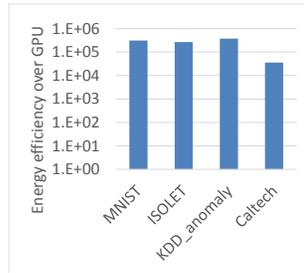
Fig. 21. Energy efficiency of the proposed system over GPU for evaluation of a new input.

Neural network utilized for KDD dataset was relatively small which made GPU execution efficient (required less off-chip memory access). As a result proposed architecture provided relatively less speedup for KDD dataset. But energy efficiency for KDD dataset is higher as it required only one neural core in the proposed system. Table III shows the number of cores used, the time and the energy for a single training data item in one iteration. Table IV shows the evaluation time and energy for a single test data. In both training and recognition phases the computation energy dominates the total system energy consumption.

Table III: For training number of cores used, the time and the energy for a single input in the proposed architecture.

| Memristor | # of core | Time (us) | Compute energy (J) | IO energy (J) | Total energy (J) |
|---|---|---|---|---|---|
| Mnist_class | 57 | 7.29 | 4.18E-07 | 8.43E-09 | 4.26E-07 |
| Isolet_class | 132 | 8.86 | 9.67E-07 | 2.66E-08 | 9.94E-07 |
| KDD_anomaly | 1 | 4.15 | 7.33E-09 | 4.47E-09 | 1.18E-08 |
| Caltech | 572 | 5.7175 | 4.19E-06 | 5.29E-08 | 4.24E-06 |

Table IV: Recognition time and energy for one input in the proposed architecture.

| Memristor | # of core | Time (us) | Compute energy (J) | IO energy (J) | Total energy (J) |
|---|---|---|---|---|---|
| Mnist_class | 57 | 0.77 | 1.42E-08 | 8.43E-09 | 2.26E-08 |
| Isolet_class | 132 | 0.77 | 3.28E-08 | 2.66E-08 | 5.94E-08 |
| KDD_anomaly | 1 | 0.77 | 2.48E-10 | 4.47E-09 | 4.73E-09 |
| Caltech | 572 | 0.77 | 1.42038E-07 | 5.29E-08 | 1.95E-07 |

## VIII. RELATED WORK

High performance computing platforms can simulate a large number of neurons, but are very expensive from a power consumption standpoint. Specialized architectures [5,6] can significantly reduce the power consumption for neural network applications and yet provide high performance.

The most recent results for memristor based neuromorphic systems can be seen in [24-26]. Alibart [32] and Preioso [33] examined only linearly separable problems. Boxun [35] demonstrates in-situ training of memristor crossbar based neural networks for nonlinear classifier designs. They proposed having two copies of the same synaptic weights, one for the forward pass and another transposed version for the backward pass. However it is practically not easy to have an exact copy of a memristor crossbar because of memristor device stocasticity. Soudry et al. [13] proposed implementation of gradient descent based training on memristor crossbar neural networks. They utilized two transistors and one memristor per synapse. Their synaptic weight precision is less than that of the two memristors per synapse neuron design utilized in this paper.

Some research efforts examined the area, power, and throughput impact of memristor based systems that carry out recognition task only [9,10]. We have examined the mapping of large deep neural networks on the multicore system where each core has limited number of inputs and outputs. Both supervised training and autoencoder based unsupervised training can be performed on the proposed architecture.

## IX. CONCLUSION

In this paper we have proposed a memristor based multicore architecture for deep network training. The system has both training and recognition (evaluation of new input) capabilities. Parallel analog operations in memristor crossbars enable efficient execution of neural networks. We developed a MATLAB framework for fast and accurate simulation of large memristor crossbars. Our experimental evaluations show that the proposed architecture can provide up to five orders of magnitude more energy efficiency over GPUs for execution of deep networks.